# Improving generation quality of pointer networks via guided attention

**Kushal Chawla[‡], Kundan Krishna[†][*], Balaji Vasan Srinivasan[‡]**

[‡]Adobe Research Big data Experience Lab, Bangalore, India
[†]Language Technologies Institute, Carnegie Mellon University, Pittsburgh, PA, USA

## Abstract

Pointer generator networks have been used successfully for abstractive summarization. Along with the capability to generate novel words, it also allows the model to copy from the input text to handle out-of-vocabulary words. In this paper, we point out two key shortcomings of the summaries generated with this framework via manual inspection, statistical analysis and human evaluation. The first shortcoming is the extractive nature of the generated summaries, since the network eventually learns to copy from the input article most of the times, affecting the abstractive nature of the generated summaries. The second shortcoming is the factual inaccuracies in the generated text despite grammatical correctness. Our analysis indicates that this arises due to incorrect attention transition between different parts of the article. We propose an initial attempt towards addressing both these shortcomings by externally appending traditional linguistic information parsed from the input text, thereby teaching networks on the structure of the underlying text. Results indicate feasibility and potential of such additional cues for improved generation.

## Introduction

Automatic text summarization (Nenkova and McKeown 2011) is the task of generating summaries of an input document while retaining the important points. These summaries are used for presenting the important content in a long piece of text in a succinct form for quick consumption of the information. Traditional methods for summarization (Nenkova and McKeown 2011) extract key sentences from the source text to construct an "extractive" summary. Various features like descriptiveness of words, word frequencies etc. have been explored to choose the sentences for summarization. However, humans summarize an article by understanding the content and paraphrasing the understood content to create the summary. Therefore, extractive summarization is unable to produce "human-like" summaries. This has led to efforts towards "abstractive" summarization which paraphrases summaries from input article content.

Early attempts at abstractive summarization created summary sentences either based on templates (Wang and Cardie 2013; Genest and Lapalme 2011) or used ILP-based sentence compression to collect parts from various sentences to generate the summary (Filippova 2010; Berg-Kirkpatrick, Gillick, and Klein 2011; Banerjee, Mitra, and Sugiyama 2015). With the advent of deep sequence to sequence models which generated text word-by-word (Sutskever, Vinyals, and Le 2014), attention based neural network models have been proposed for summarizing long sentences (Rush, Chopra, and Weston 2015; Chopra et al. 2016). Gulcehre et al. incorporated the ability to copy out-of-vocabulary words from the article to incorporate rarely seen words like names in the generated text. Tu et al. included the concept of coverage, to prevent the models from repeating the same phrases while generating a sentence. See, Liu, and Manning proposed a pointer-generator framework which incorporates these improvements, and also learns to switch between generating new words and copying words from the source article.

The pointer generator framework can efficiently handle out-of-vocabulary words and have been very successfully applied, even beyond summarization tasks (Mathews, Xie, and He 2018). However, there exists no investigations on the quality of the generated summaries from the pointer generator framework. In this work, we study the quality of the generated content and point out two key shortcomings in the quality of the generated text.

The first shortcoming is that in the quest for handling out-of-vocabulary(OOV) words by copying words from the source text, the model ends up over-compensating and learns to copy most of the times; See, Liu, and Manning observed that the mean probability of generation over copying ($p_{gen}$) to be 0.17 at test time indicating a stronger inclination towards copying. This is perhaps due to the fact that the model is incapable of differentiating between the 'necessary' parts of the input text that need to be reproduced, resulting in a summary that is largely extractive and thus, losing the desirable properties of the abstractive summaries.

The second shortcoming of this framework is the occasional presence of factual inaccuracies in the generated summary. Figure 1 shows an example incorrect summary generated by the pointer-generator network. The summary says that "*Anne Frank and her sister Margot were separated in the year* 1945" whereas the article indicates that this happened in 1944. Since the grammar in the summary is perfect, such errors are likely to be overlooked by a human annota-tor evaluating the quality of summary in the absence of the input article.

[*]work done when the author was part of Adobe Research

Figure 1: An example generation of an incorrect summary (below). The highlighted parts of the article (above) show the amount of cumulative attention that was received by each word during the entire decoding procedure.

In this paper, we present our analysis of these shortcomings in the pointer-generator framework. Our analysis demonstrates that a significant amount of errors in pointer-generator summaries arise by a combination of two effects. First, there is a tendency to add most of the words in the output summary via copying from the source article in their original sequence. Second, there is an anomaly in the transition of attention of the decoding LSTM, which controls the words being copied from the article, leading to concatenation of inconsistent clauses with each other. We refer to this as the *"shunting effect"*.

One possible reason causing these shortcomings may be the reliance of the framework on the training data as merely a sequence of words, without much interpretation of the underlying linguistic structure. This is common in several deep neural frameworks where the network is expected to 'learn' the linguistic structure from the training data. While this has been successful, we show that modifying the network to blend in additional linguistic cues enables the network to learn the structure better and thus, overcome the shortcomings.

## Pointer-generator network

The pointer generator network (See, Liu, and Manning 2017) consists of an encoder and a decoder, both based on LSTM architecture. Given an input article, the encoder takes the word embedding vectors of the source text $A = a_1 a_2 ... a_n$ and computes a sequence of encoder hidden states $h_1, h_2, .. h_n$. The final hidden state is passed to a decoder. The decoder computes a hidden state $s_t$ at each decoding time step, and an attention distribution $a^t$ is over all words,

$$a^t = \text{softmax}(e^t); e_i^t = v^T \tanh(W_h h_i + W_s s_t + b_{att}) \quad (1)$$

where $v, W_h, W_s$ and $b_{att}$ are trained model parameters. The attention model is a probability distribution over the words in the source text, which aids the decoder in generating the next word in the summary using words with higher attention. The context vector $h_t^*$ is a weighted sum of the encoder hidden states and is used to determine the next word that is to be generated, $h_t^* = \sum_{i=1}^{n} a_i^t h_i$. At each decoding time step, the decoder uses the last word $y_t$ in the summary generated so far and computes a scalar $p_{gen}$ denoting the probability of the network generating a new word from the vocabulary.

$$p_{gen} = \sigma(w_h^T h_t^* + w_s^T s_t + w_y^T y_t + b_{gen}) \quad (2)$$

where $w_h, w_s, w_y, b_{gen}$ are trained vectors. The network probabilistically decides based on $p_{gen}$, whether to generate a new word from the vocabulary or copy a word from the source text using the attention distribution. For each word $w$ in the vocabulary, the model calculates $P_{vocab}(w)$, the probability of the word getting newly generated next. $P_{vocab}$ is calculated by passing a concatenation $s_t$ and $h_t^*$ through a linear transformation with softmax activation. On the other hand, for each word $w'$ in the input article, its total attention received yields its probability of being copied. The total probability of $w$ being the next word generated in the summary, denoted by **p** is given by,

$$\mathbf{p}(w) = p_{gen} P_{vocab}(w) + (1 - p_{gen}) \sum_{i: w_i = w} a_i^t \quad (3)$$

The second term allows the framework to choose a word to copy from the input text using the attention distribution. For our experiments, we utilized the model trained using back-propagation and the Adagrad gradient descent algorithm (Duchi, Hazan, and Singer 2011).

## Generation shortcomings in pointer-generator

We observe two key shortcomings in the summaries generated by pointer-generator. The first shortcoming is the over-copying tendency exhibited. Fig. 2a shows the distribution of average $p_{gen}$ in the generated summaries. It is easy to see that around $65\%$ of the $p_{gen}$ values fall in $(0.1 - 0.2)$ bucket, indicating a strong tendency to copy. To investigate this further, we plot the percentage overlap between the summary and the input text in terms of $1-$gram, $2-$gram, $3-$gram, $4-$gram and sentences in Fig. 2b. It can be seen that $30\%$ of the sentences in the summary are reproduced as-is from the input and a mere $0.13\%$ of the 'new' words are generated from the vocabulary, indicating a strong bias towards copying as compared to generating the summary. While one may argue that this is fine if the final summary serves the purpose, this defeats the task of paraphrasing expected out of abstractive summarization.

The second shortcoming refers to the factual inaccuracies in the generated summaries. While a factual error can be caused in multiple ways, there are a few patterns that we have observed. Figure 3 illustrates three common types of errors with examples. The summary in Figure 3a incorrectly claims that a dog created a Facebook page due to *an incorrect dangling pronoun reference*. The second kind of error is caused when a sentence is *truncated prematurely* as demonstrated in Figure 3b where the summary suggests that people were advised to stop eating all products of "blue bell" whereas the article tells that only certain products originating from a particular plant are to be avoided. Premature truncation often leads to grammatically incomplete sentences as

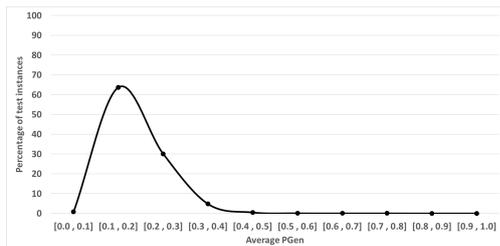

(a) Distribution of $p_{gen}$ across the generation of test summaries in CNN-Daily Mail

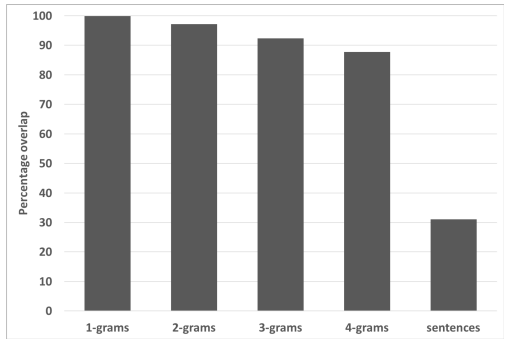

(b) Percentage overlap of $n-$grams and sentences between the generated summaries and the input text

Figure 2: Analysis of the abstractive nature of summaries generated by Pointer Generator network

per our inspections. In Figure 3c, the summary claims that a video game was developed in Japan's corridors of power. This error is caused because the system *merges words from distant clauses* without judging the compatibility.

In order to address these shortcomings, we propose to blend linguistic cues from traditional parsing such as part-of-speech tags, named entity recognition tags and co-referencing resolution with the word sequences, in order to teach the network to identify salient and connected part of the input. This way, the network will have an enhanced understanding of the input text and thus, would be able to generate better summaries.

## Making pointer generator more abstractive

To make the model copy only when necessary, we propose to add word-level *linguistic cues* such as Part-of-Speech tags, Named Entity Recognition tags and word polarity information to teach the network which words in the input are worthy of copying. For instance, proper nouns like names of persons and locations are infrequent and might often not be part of vocabulary and thus, must be copied.

In order to capture the word-level linguistic cues, we define an additional input representation $r_i$ for each word $a_i$ in the article $A$. Similar to attention, these inputs are used to define a "do-not-copy-me" distribution which gives a higher weight to the words which should not be copied. Our cues comprise of word-level attributes such as **Part-of-Speech(POS) tags** (using one-hot vectors over 4 tags, Noun, Verb, Adjective, Other) and **Named Entity Recogni-**

**Article**: ( cnn ) never mind cats having nine lives . a stray pooch in washington state has used up at least three of her own after being hit by a car , apparently whacked on the head with a hammer in a misguided mercy killing and then buried in a field – only to survive . (...) the veterinary hospital 's good samaritan fund committee awarded some money to help pay for the dog 's treatment , but mellado has set up a fundraising page to help meet the remaining cost of the dog 's care . she 's also created a facebook page to keep supporters updated . (...)
**Summary**: a stray pooch in washington state has used up at least three of her own after being hit by a car . the dog 's brush with death did not leave her unscathed . she 's also created a facebook page to keep supporters updated .

(a) Pronoun referring to an incorrect entity

**Article**:( cnn ) blue bell ice cream has temporarily shut down one of its manufacturing plants over the discovery of listeria contamination in a serving of ice cream originating from that plant . public health officials warned consumers friday not to eat any blue bell-branded products made at the company 's broken arrow , oklahoma , plant . that includes 3-ounce servings of blue bell ice cream from this plant that went to institutions in containers marked with the letters o , p , q , r , s or t behind the coding date . (...)
**Summary**:public health officials warned consumers friday not to eat any blue bell-branded products . the company is shutting down the broken arrow facility " out of an abundance of caution " to search for a possible cause of contamination . it is the third time blue bell has taken action in light of a listeria outbreak .

(b) Premature truncation leading to an incomplete sentence

**Article**: ( cnn ) the classic video game " space invaders " was developed in japan back in the late 1970 's – and now their real-life counterparts are the topic of an earnest political discussion in japan 's corridors of power . luckily , japanese can sleep soundly in their beds tonight as the government 's top military official earnestly revealed that (...)
**Summary**: the classic video game " space invaders " was developed in japan 's corridors of power . luckily , japanese can sleep soundly in their beds tonight as the government 's top military official earnestly revealed that the country 's air self defense force detects indications of an unidentified flying object that could violate our country 's airspace , it scrambles fighter jets if necessary and makes visual observation .

(c) Merging incompatible clauses to make an incorrect sentence

Figure 3: Examples of different kinds of errors in the generated summaries

**tion(NER) labels** using one-hot vectors over 12 labels - *Person, Location, Organization, Duration, Date, Cardinal, Percent, Money, Measure, Facility, GPE and Other*. This information is concatenated with the embeddings to yield a single representation for each word. Enriched word representations would enable the network to identify and seggregate the factual parts in the input article that must be "copied" to the target summary. The "**d**o-**n**ot-**c**opy-**m**e" distribution is defined as

$$I = \text{softmax}(g), \text{where,} g_i = u^T \tanh(W_{dncm} r_i + b_{dncm}).$$
(4)

$u$, $W_{dncm}$ and $b_{dncm}$ are trainable parameters. In order to in-

culcate this information into the decoding process, we compute a fixed size representation $h_{dncm}^*$ by taking a weighted average over encoder hidden states, $h_{dncm}^* = \sum_{i=1}^{n} I_i h_i$. We modify the probability of generating a new word in Eq. 2 by including this additional information:

$$p'_{gen} = \sigma(w_h^T h_t^* + w_{dncm}^T h_{dncm}^* + w_s^T s_t + w_y^T y_t + b_{gen}), \quad (5)$$

where $w_h, w_{dncm}, w_s, w_y, b_{gen}$ are trainable parameters. The model thus computes the probability by incorporating $I$ along with the attention information. The computation of total probability $\mathbf{p}(w)$ (Eq. 3) now uses this modified probability of generation:

$$\mathbf{p}(w) = p'_{gen} P_{vocab}(w) + (1 - p'_{gen}) \sum_{i:w_i = w} a_i^t \quad (6)$$

To penalize the network for paying attention to the unnecessary words, we introduce an 'insignificance' loss term, $L_t = \sum_{i=1}^{n} \min(a_i^t, I_i)$. Since we still copy words using the attention distribution, $L_t$ restricts the model to only copy the words allowed by the "do-not-copy-me' distribution $I$. This enables the model to "copy only when necessary". Finally, we train by appending this loss to the pointer generator loss comprising the negative log likelihood of the word distributions. The final loss is given by,

$$\text{loss}_t = -\log \mathbf{p}(w_t^*) + \mu L_t, \quad (7)$$

where $w_t^*$ is the target word for timestep $t$ and $\mu$ is a hyperparameter. Table 1 shows the summaries generated by different models on a test sample from CNN/Daily Mail dataset (Hermann et al. 2015; Nallapati et al. 2016). Using linguistic cues over the pointer generator baseline model results in novel sentence constructs.

We use the pre-trained pointer generator model as a starting point for our model along with the standard CNN/DailyMail dataset (Hermann et al. 2015; Nallapati et al. 2016). This dataset contains $287,226$ training instances, $13,368$ for validation and $11,490$ test instances. We further train the pretrained model on the same dataset along with the linguistic cues. We retain the same hyperparameters from the pre-trained model: 256 hidden dimensions, embedding size as 128, vocabulary size as $50,000$, 400 maximum encoding steps and 100 maximum decoding steps. At the time of testing, we decode for a maximum of 120 steps.

The proposed modification enables the model to favor the generation of novel words in the summary. We use the percentage of novel $n$-grams generated by the model as a measure of novelty in generation and report the percentage novel 1-grams, 2-grams, 3-grams, 4-grams and sentences generated. Additionally, we also report average probability of generation $p_{gen}$ during test time to further measure the effectiveness of our method. To ensure we do not sacrifice the quality of the generated summaries, we also report the ROUGE-1, ROUGE-2 and ROUGE-$L$ scores.

---

https://github.com/abisee/pointer-generator

| |
|---|
| **Input Article:** cristiano ronaldo returned to top form last weekend with five goals against granada , and he is now urging his supporters to get themselves into shape . the real madrid forward has not been at his best this year , but turned things around with a stunning display on sunday , during real madrid 's 9-1 win . and the portuguese star , and world player of the year , took to twitter to share his celebrations with his many followers , posing with a bike and encouraging them to start riding . cristiano ronaldo scored five goals as real madrid thrashed granada 9-1 on sunday afternoon . ronaldo is one of the fittest athletes in the world , and tweeted for his many fans to join him in using excercise as a way of feeling better . ' exercise all you can ! ' he wrote on twitter . ' it 's good for your body and your mind ! ' the portuguese superstar was simply irresistible as real ran riot at the bernabeu ." |
| **Pointer Generator:** cristiano ronaldo scored five goals as real madrid thrashed granada 9-1 on sunday afternoon . ronaldo is one of the fittest athletes in the world . he is now urging his supporters to get themselves into shape . it 'sgood for your body and your mind ! |
| **Pointer Generator + linguistic cues** cristiano ronaldo *scores* five goals as real madrid *beat* granada 9-1 . cristiano ronaldo returned to top form last weekend with five goals against granada . ronaldo is one of the fittest athletes in the world . |

Table 1: Summaries generated by different models on a testing instance from CNN/Daily Mail dataset. Novel words are shown in red and novel sentence constructs are in italics.

Table 2 shows the performance of the proposed approach against the baseline pointer generator network. Appending the POS and NER tag information together performed the best across all compared settings. Our best model shows almost a $50\%$ increase over baseline model in its tendency to generate novel words. To further test the statistical significance of our improvements in novelty, we employ the Fisher-Pittman permutation test (Fisher 1935) to compare the paired difference in novelty values between the baseline and our best approach. The tests indicated that our improvements in novelty are statistically significant with a p-value $< 0.05$ for 2-grams and $< 0.01$ for 3-grams, 4-grams and sentences.

### Shunting and de-shunting pointer generator

To investigate the factual inaccuracies in the generated content, we randomly select 100 articles from the test set of the CNN-Dailymail dataset and generate their summaries using the pre-trained pointer-generator model. For each article, we show it's summary along with the original article to 3 different human annotators and ask them to judge their correctness. Out of the 100 summaries, 8 were annotated to contain factual errors by a majority of raters. The annotators also reported that 22 summaries contained incomplete sentences. When asked to rate the summaries on a scale of 1(worst) to 3(best), the average score that was awarded came out to be 2.183. While the generation is mostly accurate, there are cases where there are various inaccuracies.

These factual inaccuracies are unique artifacts of the way pointer-generator networks work, and can be attributed to the shunting effect that occurs when the attention of the de-coder shifts from the middle of one sentence of the article to some other sentence, usually when both of them share a common word or multi-word phrase. This often leads to successful compression of information to create good abstractive summaries. However, the shunting effect can be erroneous if the two parts of the different sentences talk about different things.

|  | ROUGE | | | % novel n-grams | | | | | Average $p_{gen}$ |
| --- | --- | --- | --- | --- | --- | --- | --- | --- | --- |
|  | 1 | 2 | L | n=1 | n=2 | n=3 | n=4 | sentences |  |
| Baseline | 0.3577 | 0.1528 | 0.3254 | 0.13 | 2.82 | 7.67 | 12.25 | 68.99 | 0.1853 |
| Baseline + POS | 0.3519 | 0.1491 | 0.3205 | 0.10 | 3.11 | 8.43 | 13.49 | 71.88 | 0.2384 |
| Baseline + NER | 0.3482 | 0.1462 | 0.3174 | 0.11 | 3.56 | 9.32 | 14.73 | 74.20 | 0.2151 |
| Baseline + POS + NER | 0.3446 | 0.1471 | 0.3147 | 0.21 | 4.63 | 11.04 | 16.75 | 82.77 | 0.2740 |

Table 2: Performance of our proposed models on ROUGE scores, % of novel n-grams and average $p_{gen}$ against the baseline pointer generator (See, Liu, and Manning 2017)

For example, consider the articles shown in Figure 4. The highlighted parts show the amount of cumulative attention that was received by each word during the entire decoding procedure. The summary created is also shown. In the first example, it can be seen that the network starts copying the words *"in louisville, kentucky, sen. rand paul..."*, but after copying the comma, it jumps off to another sentence *"in ferguson, missouri, the shadow of michael brown..."*. In effect, the summary conveys that the shooting and protests happened in *kentucky* which is not correct. In the second example, the summary suggests that a singer's famous song *"the thrill is gone"* was in collaboration with another artist called *u2*, whereas the article says that the collaboratively produced song was *"when love comes to town"*. Here again, the shunting is caused at the closing quotes that occur in both sentences.

Such concatenations are seen in several summaries generated by the network. One possible reason can stem from Equation 1, where values of $e_i$ define the attention received by different words, since attention is a softmax activation over $e_i$. In Equation 1, the only component that changes across different decoding timesteps is $s_t$, which can be seen as a time-varying bias-shift that is added to a projection of each term's encoding, given by $W_h h_i$. Hence, if $h_i = h_j$ for words at indices $i$ and $j$, they are bound to get equal attention at all timesteps irrespective of any other factors. If the LSTM encoder encodes the same words occurring at different positions in the input into very similar encodings, then such a phenomenon can be expected to happen. This suggests that the contextual information is sometimes ignored by the LSTM encoder in which case the value of $h_i$ may just depend on the word occurring at the $i^{th}$ position.

To address the shunting effect, we propose an approach to regulate the transition of attention between decoding timesteps for maintaining factual correctness in the summary. Like in the last section, we introduce this regulation by informing the model via traditional linguistic features extracted from text injected into the encoding. Our method calculates a transition affinity function $t(i, j)$ which is higher if the transition from word index $i$ to word index $j$ is more likely to retain factual consistency. We modeled the transition affinity function using entity co-reference. By ensuring the attention stays focused on a particular entity, the method would avoid mixing up information about different entities in a sentence like in Figure 3c and also avoid dangling pronouns as in Figure 3b.

We first extract the co-reference mentions of all entities in the input article and assign each set of tokens referring to an entity with a different tag such that the same tag is used for all mentions of an entity. Thereby, the tags are also assigned to words neighboring each of the mentions. This is done by extracting the smallest subsequence of words around the mention which form a complete clause. For this we parse a sentence and then select a subtree which contains the mention and has the root non-terminal signifies a clause. The transition affinity function $t(i, j)$ is defined here to be the number of tags that occur both on the $i^{th}$ and the $j^{th}$ word.

Since factual errors caused via the shunting effect are due to inconsistent attention transition, we bias the value of attention for each word based on its transition affinities with the words that received high attention in the previous timestep. We do this by changing the calculation of $e_i$ such that,

$$e_i^t = v^T \tanh(W_h h_i + W_s s_t + b_{att}) + W_a \sum_{j=1}^{n} a_j^{t-1} t(j, i)$$
(8)

$W_a$ is an additional scalar parameter which is learned here.

During the final training iterations, we also use an auxiliary loss function to incorporate transition affinity. We calculate the average transition affinity $\sum_{m=1}^{n} a_j^{t-1} t(j, i)$ over all words $w_i$ and all decoding timesteps. To maximize this average value, the negative of this is appended to the loss function for the optimizer like in Eq. 7.

In the absence of existing metrics to measure the factual correctness, we conduct human evaluation on the same 100 articles used before. The model based on the proposed modification was rated better on factual correctness by the annotators for 31 summaries, including all but 1 summary that had factually inaccurate baseline summaries, according to a majority of raters. This suggests that our model is able to avoid most of the errors that were committed by baseline pointer-generator network. Due to the relatively limited number of cases of such factual inaccuracies, the change in ROUGE between the various setups was negligible and hence we have not reported here.

## Conclusion & Future work

Detecting and fixing the different kinds of errors occurring in abstractive summarization systems is a fertile area for research. While there has been work to remove errors

**Article**
-lrb- cnn -rrb- the nation 's top stories will be unfolding tuesday in __courthouses__ and political arenas across the country . massachusetts is hosting two of the highest-profile court trials in recent memory -- those of former new england patriot aaron hernandez and boston bombing suspect dzhokhar tsarnaev . both lengthy trials are coming to a close . in louisville , kentucky , sen. rand paul made the __not-so-surprising__ announcement that he will run for president , while in chicago , voters will head to the polls in a very surprising runoff between mayor rahm emanuel and challenger jesus `` __chuy__ '' garcia . and in ferguson , missouri , the shadow of michael brown and the protests over his shooting by officer darren wilson will loom large over the city 's

**Summary**
massachusetts is hosting two of the highest-profile court trials in recent memory . both lengthy trials are coming to a close . in louisville , kentucky , the shadow of michael brown and the protests over his shooting .

(a) Merging incompatible clauses to make an incorrect sentence

**Article**
in the 1980s , gibson officially dropped the model number on the guitar he used last and most . it became a custom-made signature model named lucille , manufactured exclusively for the `` king of the blues . '' some of his hits include `` the thrill is gone , '' which won him his first grammy in 1970 , `` there must be a better world somewhere '' and `` when love comes to town , '' a collaboration with u2 . last year , the __bluesman__ suffered from dehydration and exhaustion after a show in chicago , forcing him to cancel the remainder of his tour . cnn 's greg botelho and sonya __hamasaki__ contributed to this report .

**Summary**
b.b. king 's dehydration was caused by his type ii diabetes . he was inducted into the rock and roll hall of fame in 1987 . some of his hits include `` the thrill is gone , '' a collaboration with u2 .

(b) Pronoun referring to an incorrect entity.

Figure 4: Errors due to the shunting effect (out-of-vocabulary words are decorated like __this__).The highlighted parts show the amount of cumulative attention that was received by each word during the entire decoding procedure.

in extractive summarization (Durrett, Berg-Kirkpatrick, and Klein 2016), there haven't been enough efforts given to design such improvements for abstractive summarization. The lack of explainability in deep learning based models makes it even more necessary.

In this paper, we point out two key shortcomings of the pointer generator framework and address it via additional linguistic cues from traditional parsing. The resulting solution is promising and illustrates the need to investigate the use of traditional linguistic information towards enhancing the understanding of deep learning models and thus, enabling better generation.